\documentclass{article}
\usepackage{amsmath,epsfig}
\usepackage[preprint]{spconfa4}

\let\OLDthebibliography\thebibliography
\renewcommand\thebibliography[1]{
  \OLDthebibliography{#1}
  \setlength{\parskip}{0pt}
  \setlength{\itemsep}{0pt plus 0.3ex}
}

\usepackage{multicol,multirow,color}
\usepackage[pagebackref=false,breaklinks=false,bookmarks=false]{hyperref}
\usepackage{footnote}

\begin{document}\sloppy
\def\x{{\mathbf x}}
\def\L{{\cal L}}

\newcommand{\ru}{\rule{0mm}{3mm}}

\title{Are GAN generated images easy to detect?  A critical analysis of the state-of-the-art}
\name{D. Gragnaniello, D. Cozzolino, F. Marra, G. Poggi and L. Verdoliva}
\address{University Federico II of Naples}
\maketitle

\begin{abstract}
The advent of deep learning has brought a significant improvement in the quality of generated media. However, with the increased level of photorealism, synthetic media are becoming hardly distinguishable from real ones, raising serious concerns about the spread of fake or manipulated information over the Internet. In this context, it is important to develop automated tools to reliably and timely detect synthetic media. 
In this work, we analyze the state-of-the-art methods for the detection of synthetic images, highlighting the key ingredients of the most successful approaches, and comparing their performance over existing generative architectures. We will devote special attention to realistic and challenging scenarios, like media uploaded on social networks or generated by new and unseen architectures, analyzing the impact of suitable augmentation and training strategies on the detectors' generalization ability.\end{abstract}
\section{Introduction}
\label{sec:intro}

\begin{figure}[t!]
    \centering
    \includegraphics[width=0.95\linewidth]{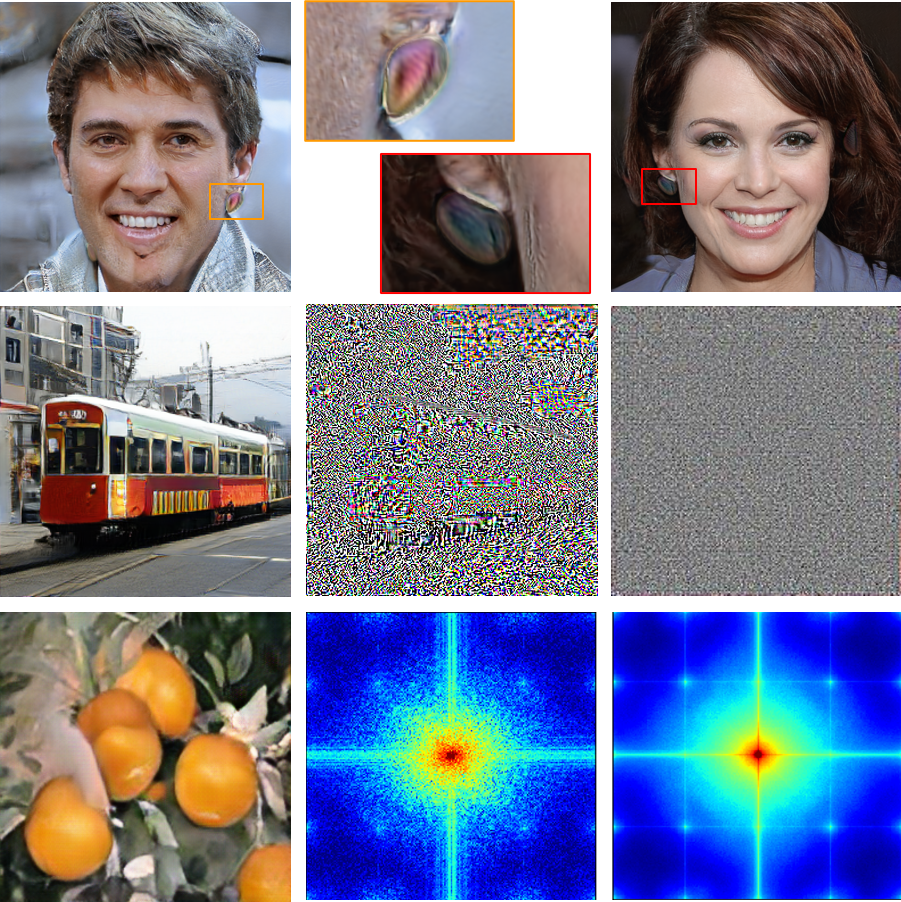}
    \caption{Examples of GAN synthetic images together with their visible and not visible artifacts. From top to bottom: color artifacts, artificial fingerprint and its averaged version, Fourier spectrum and its averaged version.}
    \label{fig:synthetic_images}
\end{figure}

In recent years, there has been intense research on the generation of synthetic media, and a large number of deep learning-based methods have been proposed to this end.
Generative adversarial networks (GAN), in particular, have brought tremendous quality improvements.
There are GAN-based methods to generate images from scratch as well as to modify the attributes of an existing image.
A number of exciting applications exist already.
However, this technology can also be used for malicious purposes, for example to generate fake profiles on social network or to generate fake news.
Even the most careful observer can now be fooled by GAN-generated images, not to mention the average Internet user.
Therefore, there is urgent need for automatic tools that can reliably distinguish real content from manipulated content.

Indeed, despite their high visual quality, synthetic images bear peculiar traces left by the generation process that can be exploited to detect them.
Sometimes they present visible artifacts, such as color anomalies or lack of symmetries, see Fig.\ref{fig:synthetic_images}.
Nonetheless, with the fast pace of technology, these obvious imperfections will likely disappear soon.
More solid and lasting evidence, though, are the invisible artifacts closely linked to the architecture of the generative network.
Indeed, GAN-generated images have been shown \cite{Marra2019DoGAN,Yu2019} to incorporate regular patterns, a sort of artificial fingerprints, specific to each individual GAN architecture, see again Fig. \ref{fig:synthetic_images}.
Such patterns also show themselves as peaks in the Fourier domain, not present in the spectral distribution of natural images.
They clearly depend on the up-sampling operations typical of each GAN architecture.
However, networks with the same architecture, but trained on different data, also have different fingerprints, thus calling for more accurate explanations.

Recently, several GAN-image detectors have been proposed in the literature, some using explicitly the features described above, others relying entirely on suitably trained deep networks \cite{Verdoliva2020media}.
In general, they seem to convey the idea that detecting GAN images is not really a challenging task.
However, we believe this is an overly optimistic view.
In our experience, detection performance impairs dramatically as soon as some favorable conditions disappear.
More specifically, both the lack of well aligned training data and the presence of significant image distortions affect performance heavily.
Unfortunately, these situations are way too common on the web.
New generators are proposed by the day, for which training examples are not available, and images are routinely compressed and resized, destroying precious evidence.

In this paper we carry out a systematic experimental study with the aim to establish where we really are in GAN-image detection.
Towards this end, several of the most promising detectors are tested, 
considering challenging yet realistic scenarios, a number of different datasets, and using performance metrics that are appropriate for large-scale screening.
Besides providing a solid reference for future proposals,
the comparative analysis of results allows us to single out some key features of successful solutions, clearing the way for the design of new and more effective tools.
In the rest of the paper we first describe state-of-the-art approaches, then we present the datasets used to carry out our experimental analysis and finally draw conclusions.

\section{State-of-the-art methods}
\label{sec:detectors}

\subsection{Learning spatial domain features}

Early works on GAN image forensics focus on the attribution problem, trying to identify the image provenance.
In fact, just like real cameras, which mark each acquired image with a device-dependent signature, also GAN architectures insert in each generated image a sort of fingerprint.
This latter depends not only on the specific model, but also on the dataset used for training \cite{Marra2019DoGAN, Yu2019},  thus enabling model identification.

Other methods exploit the intrinsic constraints of GAN generators.
For example, \cite{McCloskey2019} leverages the fact that GANs produce only a limited range of intensity values, and do not generate saturated and/or under-exposed regions.
Likewise, \cite{Li2020} exploits GANs failure to accurately preserve the natural correlation among color bands.
Consequently, to extract discriminative features for detection, the chrominance components are high-pass filtered and summarized by their co-occurrence matrices.
It is worth noting that co-occurrences of high-pass filtered versions of the image are popular tools in image forensics 
since invisible artifacts are often present in the high-frequency signal components \cite{Verdoliva2020media}.
In fact, co-occurrence matrices extracted from the RGB channels are also used in \cite{nataraj2019detecting} as input of a CNN.
and in \cite{Barni2020CNN} across color bands.

A first investigation of detectors based on very deep networks is carried out in \cite{Marra2018},
where state-of-the-art pre-trained CNNs, like Xception, Inception, and DenseNet, are shown to ensure excellent performance for GAN image detection. 
In particular, they turn out to outperform CNN models specifically tailored to forensics tasks and trained from scratch, and especially in the most challenging scenarios. 

\subsection{Learning frequency domain features}

GAN images display clear traces of their synthetic origin in the Fourier domain.
The detector proposed in \cite{Zhang2019} exploits the presence of spectral peaks caused by the upsampling operations routinely performed in most GAN architectures.
A frequency-domain analysis is carried out also in \cite{frank2020leveraging} to study the presence of artifacts across different network architectures, datasets and resolutions.
Again, these artifacts are used to tell apart generated images from real ones.
In particular, a CNN-based classifier is trained with Fourier spectra taken from both real images and their synthetic versions obtained trough an adversarial autoencoder.
Likewise, in \cite{Durall2020} it is shown that GAN images do not faithfully mimic the spectral distributions of natural images.
A simple detector is proposed that takes the energy spectral distribution as input feature.
The authors also propose a spectral loss to use during GAN training so as to limit the appearance of spectral artifacts.

\subsection{Learning features that generalize}

The fully supervised approaches described above are all very effective when the GAN images under test come from a model that is also present in training.
However, they fail to generalize to data generated by new unseen models.
Therefore, some methods have been proposed recently to address this problem.
In \cite{Cozzolino2018, Du2019} few-shot learning strategies are proposed, with an autoencoder-based architecture, to adapt to new manipulations with just a few examples.
In \cite{Marra2019incremental}, instead, an approach based on incremental learning is used.
Despite the improved generalization, these methods still need some examples of the new GAN architecture, which is not always realistic.

A different solution is proposed in \cite{xuan2019generalization}. 
The idea is to carry out augmentation by gaussian blurring so as to force the discriminator to learn more general features. 
A similar approach is followed in \cite{Wang2020} where a standard pre-trained model, ResNet50, is further trained with a strong augmentation based on compression and blurring. 
Experiments show that, even by training on a single GAN architecture, the learned features generalize well to unseen architectures, datasets, and training methods.
A different perspective is taken in \cite{chai2020makes} where a fully-convolutional patch-based classifier is proposed. The authors show that by focusing on local patches rather than global structure,
they can achieve better performance.

\setlength{\tabcolsep}{2pt}
\begin{table}[t!]
    \centering
    \begin{tabular}{llc}
    \hline
    \multicolumn{3}{c}{\bf Low Resolution ($256\times256$)} \\
    Name & Content & \# Images \\ \hline
    Various                              & ImageNet, COCO, Unpaired-real                      & 11.1k\\
    StyleGAN      & Generated objects (LSUN)                            &  6.0k \\
    StyleGAN2     & Generated objects (LSUN)                            &  8.0k \\
    BigGAN        & Generated objects (ImageNet)                        &  2.0k \\
    CycleGAN      & Image-to-image translation                          &  4.0k \\
    StarGAN       & Generated faces   (CelebA)                          &  2.0k \\
    RelGAN        & Generated faces   (CelebA)                          &  3.0k \\
    GauGAN        & Generated scenes  (COCO)                            &  5.0k \\ \hline \hline
    \multicolumn{3}{c}{\bf High Resolution ($1024\times1024$)} \\
    Name & Content & \# Images \\\hline
    RAISE~\cite{dang2015raise}    		 & Central crop of real photos                         &  7.8k \\
    ProGAN             & Generated faces   (CelebA-HQ)                       &  3.0k \\
    StyleGAN           & Generated faces   (CelebA-HQ)                       &  3.0k \\
    StyleGAN           & Generated faces   (FFHQ)                            &  3.0k \\
    StyleGAN2          & Generated faces   (FFHQ)                            &  3.0k \\\hline
    \end{tabular}
    \caption{Datasets used for testing the methods under analysis.}
    \label{tab:datasets}
\end{table}

\section{Datasets}

In our experimental analysis all networks are trained and tested on the very same data.
For training, we use the dataset provided by \cite{Wang2020},
comprising 362K real images extracted from the LSUN dataset and 362K generated images obtained by 20 ProGAN~\cite{karras2018progressive} models, each trained on a different LSUN object category.
All images have a resolution of 256$\times$256 pixel.
A subset of 4K images is used for validation.

Since our main aim is to verify the model transferability, in the testing phase we use images coming from GAN architectures never seen in training.
Testing datasets are listed in Table \ref{tab:datasets}.
They include both low resolution (256$\times$256) and high resolution (1024$\times$1024) images, generated by:
StyleGAN~\cite{karras2019style},
StyleGAN2~\cite{karras2020analyzing},
BigGAN~\cite{brock2018large},
CycleGAN~\cite{zhu2017unpaired},
StarGAN~\cite{choi2018stargan},
RelGAN~\cite{wu2019relgan}, and
GauGAN~\cite{park2019SPADE}.
We exclude low resolution ProGAN images, since they are used for training, but consider instead their high-resolution versions.
Overall, we have about 39K synthetic images.
Then we have 11.1K low-resolution real images coming from ImageNet, COCO~\cite{lin2014microsoft}, and Unpaired real dataset~\cite{zhu2017unpaired},
and 7.8K high-resolution images, extracted from the RAISE~\cite{dang2015raise} dataset.
We do not use high-resolution images from the CelebA-HQ dataset, as done elsewhere, since they are GAN-upsampled versions of the low-resolution real images.

{\small
\setlength{\tabcolsep}{2mm}
\begin{table*}
\centering
\begin{tabular}{llll} \hline
\ru   \textbf{Ref.}                 & \textbf{Acronym} & \textbf{Description}                                          & \textbf{Test strategy}             \\ \hline  
\ru   \cite{Marra2018}              & Xception         & pre-trained Xception without augmentation                     & no cropping and no resizing        \\ 
\ru   \cite{Boroumand2019deep}      & SRNet            & 12-layer network with no down-sampling in the first 7 layers  & no cropping and no resizing        \\ 
\ru   \cite{Zhang2019}              & Spec             & pre-trained ResNet34 with image spectrum as input             & central cropping ($224\times 224$) \\ 
\ru   \cite{xuan2019generalization} & M-Gb             & 6-layer network with gaussian blurring augmentation           & resizing ($128\times 128$)         \\ 
\ru   \cite{nataraj2019detecting}   & Co-Net           & 8-layer network with co-occurrence matrix as input            & no cropping and no resizing        \\ 
\ru   \cite{Wang2020}               & Wang2020         & pre-trained ResNet50 with blurring and compression augment.   & no cropping and no resizing        \\ 
\ru   \cite{chai2020makes}          & PatchForensics   & first blocks of Xception trained at patch-level               & resizing ($299\times 299$)         \\ \hline 
\end{tabular}
\caption{List of the methods used in our experiments together with the test strategy, as proposed in the original papers. }
\label{table:supervised}
\end{table*}
}

\section{Experimental results}
\label{sec:results}

In our analysis we compare a number of detectors:
Xception \cite{Marra2018}, 
SRNet \cite{Boroumand2019deep}, 
Spec \cite{Zhang2019}, 
M-Gb \cite{xuan2019generalization}, 
Co-Net \cite{nataraj2019detecting}, 
Wang2020 \cite{Wang2020}, 
PatchForensics \cite{chai2020makes}.
Their main features are summarized in Table \ref{table:supervised}.
Together with methods specifically proposed for GAN image detection, and already described in Section 2, we also include SRNet, originally proposed for steganalysis.
Indeed, steganalysis and image forensics pursue very similar goals, and successful methods transfer well from one domain to the other \cite{Verdoliva2020media}.
In particular, we find SRNet worth studying because, to preserve features related to noise residual, it performs no down-sampling in the first layers of the network,
a solution of potential interest for GAN detection.

\begin{figure*}
    \centering
    \includegraphics[width=0.99\linewidth,trim=10 0 10 0,clip]{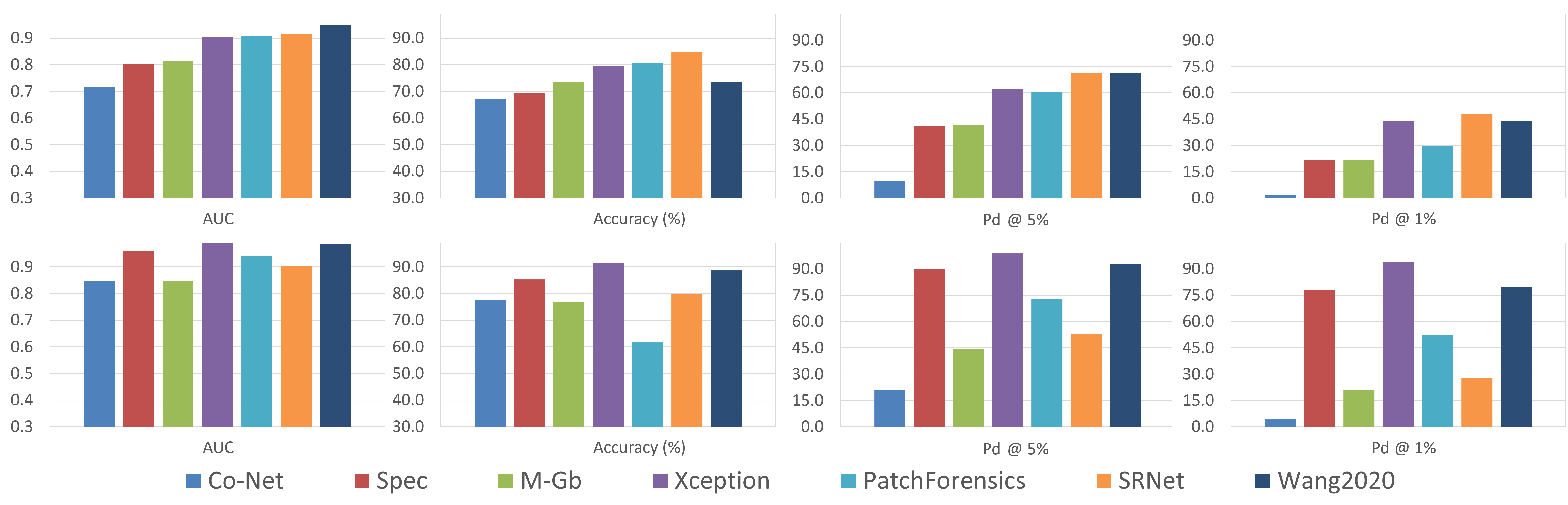}
    \caption{Results of the methods under comparison in terms of AUC, Accuracy, Pd@5\% and Pd@1\% for all the tested methods on low-resolution (top) and high resolution images (bottom).}
    \label{fig:comparisons_bar}
\end{figure*}

The first set of experiments aims at assessing the generalization ability of the tested methods.
Results are shown in Fig.\ref{fig:comparisons_bar} for low-resolution (top) and high-resolution (bottom) images in terms of several performance metrics:
area under the receiver-operating curve (AUC), accuracy at the fixed threshold of 0.5, and probability of detection for a 5\% (Pd@5\%) and 1\% (Pd@1\%) false alarm rate (FAR).
AUC results on low-resolution (LR) images are generally very good, considering that training and testing data are not aligned, with several methods exceeding the 0.9 level.
However, accuracy results are much less encouraging, since a fixed threshold is used, 
lacking the optimal one.
Considering the Pd@FAR metric, results become pretty bad, and only some methods keep ensuring a good detection ability.
On the other hand, if detectors are to be used for systematic screening of media content, only very low FARs are acceptable.
Results are somewhat better for high-resolution (HR) images but the same general behavior is observed.

\begin{figure*}
    \centering
    \includegraphics[width=0.99\linewidth]{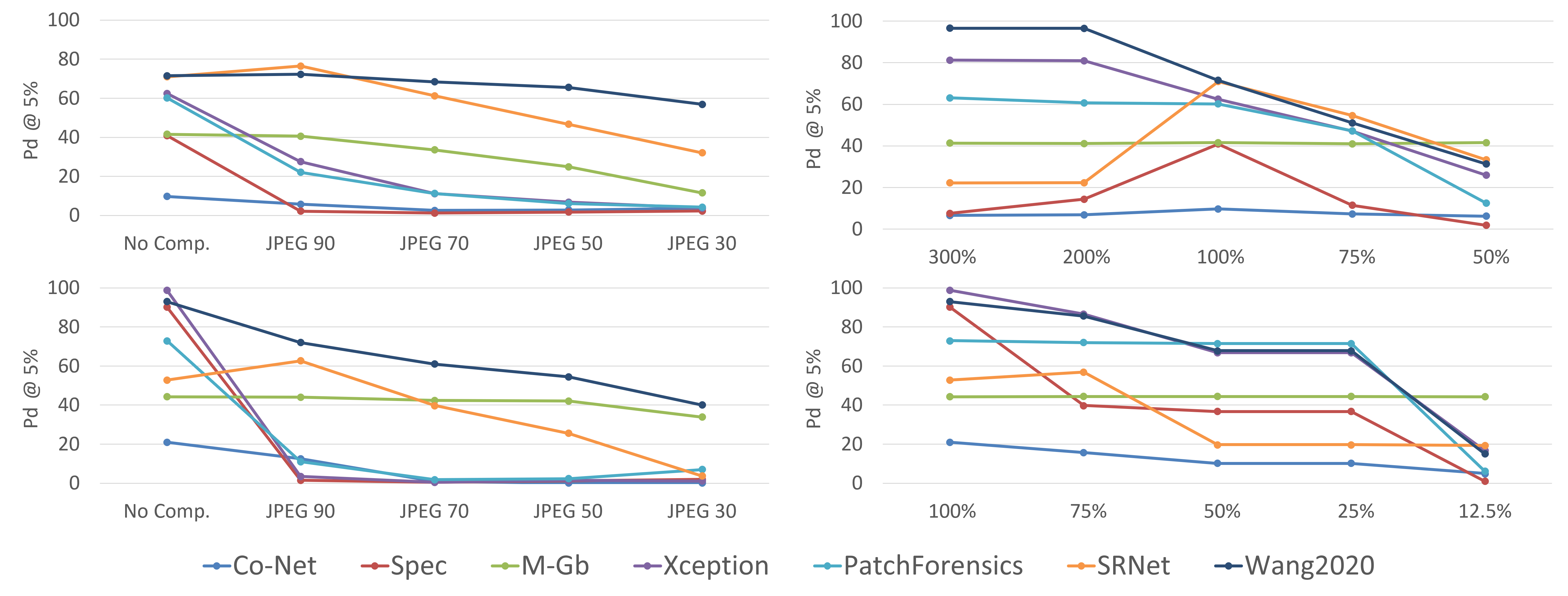}
    \caption{Results of the methods under comparison in terms of Pd@5\% as a function of JPEG compression level and resizing factor. LR images (top) are both enlarged and reduced in size, while HR images (bottom) are only reduced.}
    \label{fig:comparisons_graph}
\end{figure*}

The above results are obtained on uncompressed images at their original size.
However, on social networks images are routinely compressed and resized, so we now investigate robustness to these processing steps.
Fig.\ref{fig:comparisons_graph} reports the Pd@5\% performance for LR and HR images as a function of compression factor and resizing scale.
Several methods suffer dramatic impairments as soon as they move away from the ideal case of no compression and 100\% scale.
A relatively stable performance is ensured by methods trained with augmentation.
In any case, a 2x downsampling has catastrophic effects, as expected given the nature of GAN artifacts.

\begin{figure*}
    \centering
    \includegraphics[width=0.99\linewidth,trim=10 0 10 0,clip]{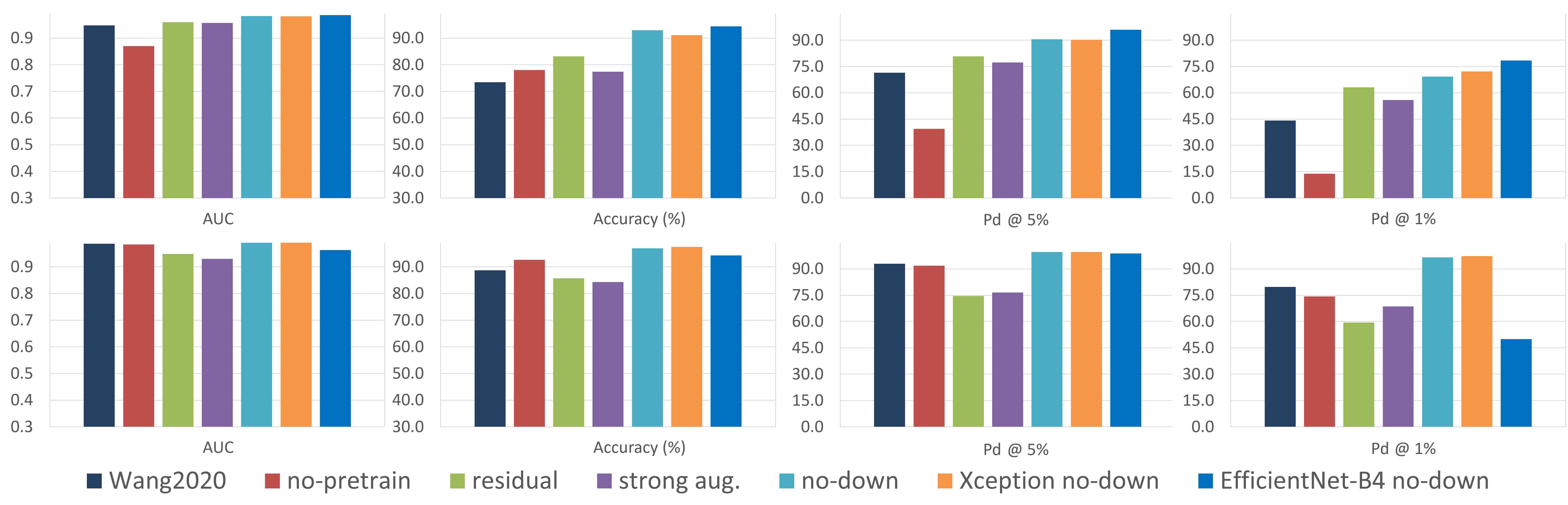}
    \caption{Results of the baseline (Wang2020) and its variants in terms of AUC, Accuracy, Pd@5\% and Pd@1\% for variants of Wang2020 on low-resolution (top) and high resolution images (bottom).}
    \label{fig:improvement_bar}
\end{figure*}

\begin{figure*}
    \centering
    \includegraphics[width=0.99\linewidth]{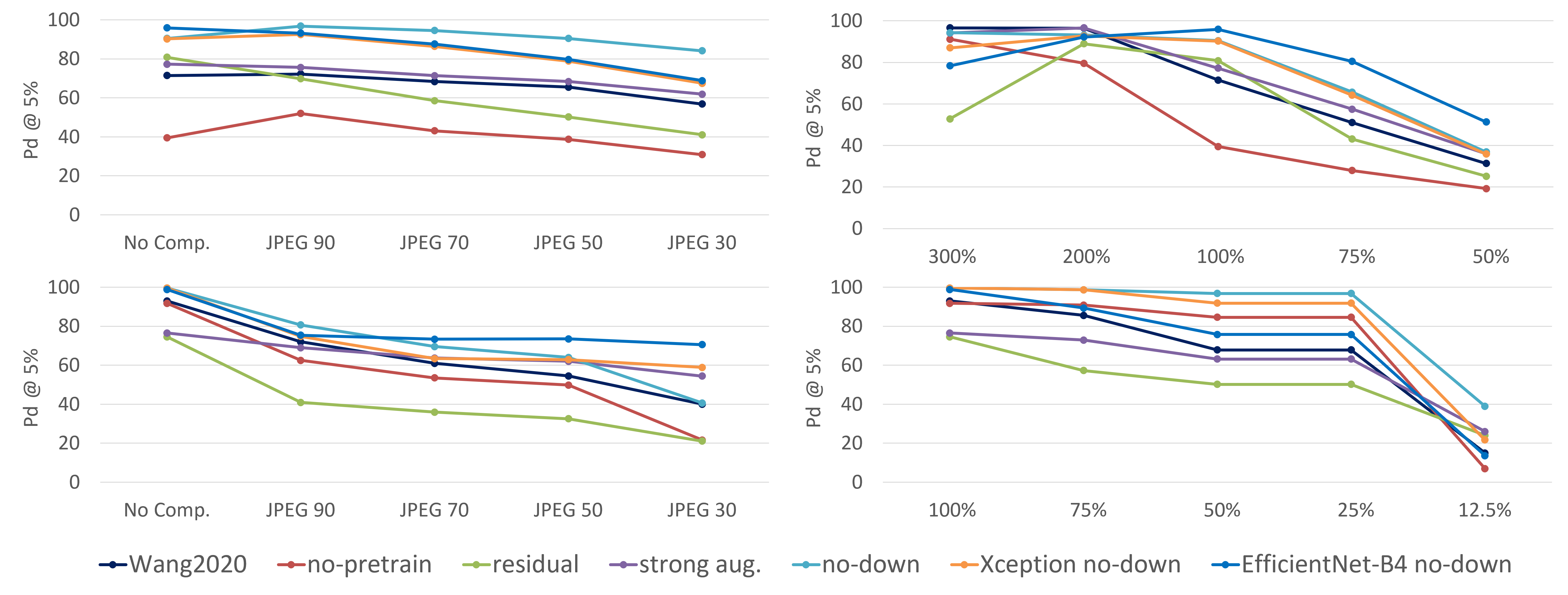}
    \caption{Results of the baseline (Wang2020) and its variants in terms of Pd@5\% as a function of JPEG compression level and resizing factor. LR images (top) are both enlarged and reduced in size, while HR images (bottom) are only reduced.}
    \label{fig:improvement_graph}
\end{figure*}

In order to move a step towards a better solution, we carry out further investigations aimed at identifying the key ingredients of the most promising solutions.
Therefore, we consider as baseline the method proposed in \cite{Wang2020}, which provided a good and stable performance in the previous experiments.
On this framework, we introduce the following variations: 
remove Imagenet pre-training (no-pretrain), 
include an initial layer for residual extraction (residual), 
do not perform down-sampling in the first layer as suggested by \cite{Boroumand2019deep} (no-down), 
perform a stronger augmentation (strong-aug) by including gaussian noise adding, geometric transformations, cut-out, and brightness and contrast changes.
In addition, for the no-down variant, we also change the backbone network, replacing ResNet50 with XceptionNet (Xception no-down) and Efficient-B4 (Efficient no-down).
Results for the various metrics are shown in Fig.\ref{fig:improvement_bar} in the absence of compression and resizing,
while Fig.\ref{fig:improvement_graph} shows results in terms of Pd@5\% as a function of compression level and scaling factor.

Finally, in Table 3 we show the results for the baseline and the best variant over all the different GAN architectures. We also considered a new version of the best variant that is trained on 23 StyleGAN2 models. Avoiding down-sampling in the first block of the architecture provides an average gain of about 15\% in terms of accuracy and 14\% in terms of Pd@5\%. Overall accuracy is always above 90\% irrespective of the type of architecture
and further improves (above 97\%) if training is carried out on StyleGAN2 \footnote{code available at: \url{https://github.com/grip-unina/GANimageDetection}}.

Although these experiments are very limited and preliminary, they provide some interesting hints for future research.
First of all, they confirm the importance of diversity to increase robustness, like ImageNet pre-training,
as already observed in steganalysis \cite{Yousfi2020Imagenet}.
In addition, they suggest there is still much room for improvements with respect to the existing solutions, especially in terms of robustness to compression and resizing.
In particular, the no-down variants appear quite promising and suggest to move along this direction by performing full-resolution end-to-end processing to design better and more robust detectors,
as also suggested in \cite{marra2020full}.

\section{Conclusions}
\label{sec:conclusions}

In this paper, we carried out a fair experimental analysis of several existing detectors, considering various challenging cases. 
Our first conclusion is that we are still very far from having reliable tools for GAN image detection.
Misalignment between training and test, compression and resizing are all sources of serious impairments and, at the same time, highly realistic scenarios for real-world applications.
On the positive side,
this analysis allows us to single out some key ingredients of successful solutions, and provides hints for future research.

\begin{table}
    \centering
    {\small
    \begin{tabular}{|cl|ccc|}
    \cline{3-5}
 \multicolumn{2}{c|}{\multirow{3}{*}{\ru Accuracy / Pd@5\%}} &  Wang2020  & Best variant & Best variant \\
 \multicolumn{2}{c|}{} & ~~~~(baseline)~~~~ & (no-down) & (no-down) \\
 \multicolumn{2}{c|}{} & [ProGAN]   &  [ProGAN]  & [StyleGAN2] \\\hline
\ru  \multirow{8}{*}{ \rotatebox{90}{Low Res.}}
    &        ProGAN  & \underline{99.3~~/100.0}  & \underline{94.7~~/100.0}  & 99.8~~/100.0 \\
\ru &      StyleGAN  & 75.9~~/~~73.9 & 93.7~~/~~93.1 & 99.9~~/100.0 \\
\ru &     StyleGAN2  & 71.5~~/~~69.0 & 92.2~~/~~88.8 & \underline{99.9~~/100.0} \\
\ru &        BigGAN  & 59.2~~/~~45.2 & 93.5~~/~~92.0 & 96.5~~/~~99.4 \\
\ru &      CycleGAN  & 77.4~~/~~80.5 & 90.3~~/~~81.5 & 96.5~~/~~99.5 \\
\ru &       StarGAN  & 84.3~~/~~89.4 & 94.5~~/~~97.6 & 99.9~~/100.0 \\
\ru &        RelGAN  & 63.6~~/~~56.0 & 92.8~~/~~86.6 & 99.7~~/100.0 \\
\ru &        GauGAN  & 82.5~~/~~86.3 & 93.6~~/~~93.5 & 90.8~~/~~97.1 \\ \hline \hline
  \multirow{4}{*}{ \rotatebox{90}{High Res.~~}}
    & ProGAN         & 99.7~~/100.0  & 97.1~~/100.0  & 99.7~~/100.0 \\
\ru & StyleGAN(Cel.) & 99.3~~/100.0  & 97.1~~/100.0  & 99.7~~/100.0 \\
\ru & StyleGAN(FFHQ) & 82.6~~/~~93.7 & 96.6~~/~~98.7 & 99.7~~/100.0 \\
\ru & StyleGAN2      & 73.2~~/~~78.1 & 96.9~~/~~99.6 & \underline{99.7~~/100.0} \\ \hline
    \end{tabular}
    }
    \caption{Accuracy and Pd@5\% for the baseline and the best variant that avoids down-sampling in the first block. First two columns training is on ProGAN, last column on StyleGAN2.}
    \label{tab:my_label}
\end{table}

\section*{Acknowledgements}
This material is based on research sponsored by the Defense Advanced Research Projects Agency (DARPA) and the Air Force Research Laboratory (AFRL) under agreement number FA8750-20-2-1004.
The views and conclusions contained herein are those of the authors and should not be interpreted as necessarily representing the official policies or endorsements, either expressed or implied, of DARPA and AFRL or the U.S. Government. In addition, this work is supported by Google and by the PREMIER project, funded by the Italian Ministry of Education, University, and Research within the PRIN 2017 program.

\bibliographystyle{IEEEbib}
\bibliography{refs}

\end{document}